\pdfoutput=1
\documentclass[11pt]{article}

\usepackage[]{acl}

\usepackage{times}
\usepackage{latexsym}
\usepackage{amsmath}
\usepackage{microtype}
\usepackage{times}
\usepackage{epsfig}
\usepackage{graphicx}
\usepackage{amsmath}
\usepackage{amssymb}

\usepackage{subcaption}
\usepackage{booktabs}
\usepackage{multirow}
\usepackage{enumitem}
\usepackage{algorithmic}
\usepackage{algorithm}
\usepackage{xcolor}
\usepackage{multirow}
\usepackage[T1]{fontenc}
\usepackage[utf8]{inputenc}

\usepackage{microtype}
\usepackage{comment}
\usepackage[normalem]{ulem}
\useunder{\uline}{\ul}{}
\newcommand{\modelname}{ REINA }

\title{Training Data is More Valuable than You Think:\\ A Simple and Effective Method by Retrieving from Training Data}

\author{Shuohang Wang, Yichong Xu, Yuwei Fang, Yang Liu, Siqi Sun,  Ruochen Xu, \\ \bf Chenguang Zhu, Michael Zeng \\
Microsoft Azure Cognitive Services Research \\
\small \{\tt shuowa, yicxu, yuwfan, yaliu10, siqisun, ruox, chezhu, nzeng\}@microsoft.com}

\begin{document}
\maketitle
\begin{abstract}
Retrieval-based methods have been shown to be effective in NLP tasks via introducing external knowledge. 
However, the indexing and retrieving of large-scale corpora bring considerable computational cost.
Surprisingly, we found that \textbf{RE}trieving from the tra\textbf{IN}ing dat\textbf{A} (\textbf{REINA}) only can lead to significant gains on multiple NLG and NLU tasks. We retrieve the labeled training instances most similar to the input text and then concatenate them with the input to feed into the model to generate the output.
Experimental results show that this simple method can achieve significantly better performance on a variety of NLU and NLG tasks, including summarization, machine translation, language modeling, and question answering tasks. For instance, our proposed method achieved state-of-the-art results on XSum, BigPatent, and CommonsenseQA. Our code is released.\footnote{ \url{https://github.com/microsoft/REINA}}
\end{abstract}

\section{Introduction}
In natural language processing, retrieval-based methods work by fetching textual information related to the input from large corpora.
The model then takes both the input and retrieved results as input to generate results.
This can often improve the performance as the model is exposed to related knowledge not present in the input.
As a result, retrieval-based methods have been successfully applied in many tasks such as open-domain question answering~\cite{chen2017reading}, language modeling~\cite{guu2018generating,khandelwal2019generalization} and machine translation~\cite{khandelwal2020nearest}. However, these methods require building an index of large-scale corpus, and the retrieval leads to a significant computational burden. For example, the kNN-MT model for machine translation has a generation speed two orders of magnitude slower than traditional MT models~\cite{khandelwal2020nearest}.

\begin{figure}[t]
  \centering
  \includegraphics[width=0.9\linewidth]{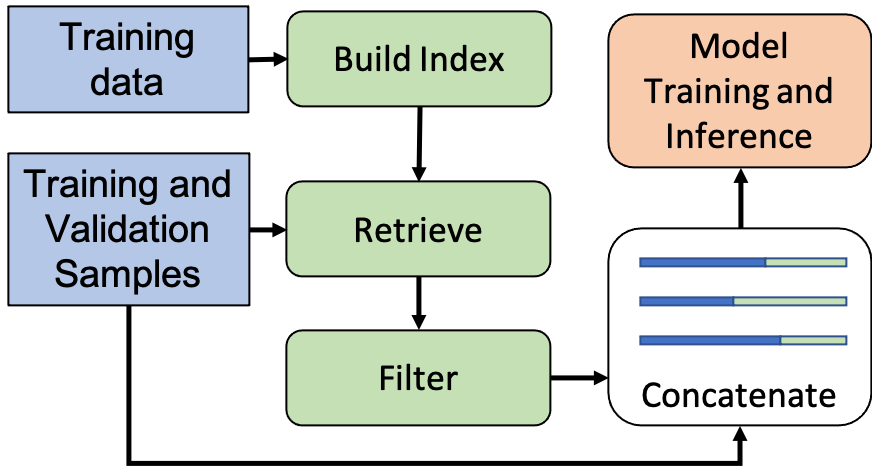}
  \caption{\modelname pipeline of model training/inference with retrieval from training data. Filter only happens at training, as the same training sample will be  retrieved from the index. For each instance, we concatenate the input with the retrieved content, i.e., data and/or labels, for model training and inference.}
  \label{fig:pipe}
\end{figure}
On the other hand, in the supervised learning setting, the text most similar in distribution to the data in inference is the training data. Thus, we explore whether retrieving from the training data, which is usually much smaller than a large-scale corpus, can help improve the performance.
Specifically, we first index a task's labeled training data as input-label pairs. 
Then, during both training and testing, we retrieve the input-label pairs most similar to the current input\footnote{During training, we exclude the training instance itself from the retrieval results to avoid data leakage.}. Finally, we concatenate the retrieved training pairs with the input and feed it into the model. An overview of our method is shown in Figure~\ref{fig:pipe}.

We note that our method is similar to recent works in prompt learning~\cite{brown2020language,liu2021makes}, where a set of labeled data is carefully chosen based on the input and then included in the prompt for few-shot learning. Our method also bears a resemblance to non-parametric instance-based learning~\cite{gu2018search}. However, a critical difference is that we focus on the supervised learning setting, where the model parameters are fine-tuned to learn from given examples to achieve much higher performance than few-shot learning or non-parametric methods.

In the experiments, we evaluate our method on four popular types of NLP tasks: summarization, language modeling, machine translation, and question answering. We find that $i$) after integrating REINA, we can achieve significantly better performance on these tasks, 11 datasets in total, than models with different pre-trained models; $ii$) \modelname leads to SOTA performance on the datasets of XSum, CommonsenseQA (Leaderboard No.1), and BigPatent; 
$iii$) \modelname can scale up more easily by leveraging more labeled data from other datasets via retrieval, outperforming baselines which is trained on the same set of data.
$iv$) the results on 3 summarization tasks show that BART-base with \modelname rivals BART-large, which contains twice more parameters now.

The effectiveness of our approach on summarization tasks provides insights into the core of supervised learning. Even with hundreds of millions of parameters, a model cannot memorize all the patterns in the training data. Thus, recapturing related training data as a side-by-side reminder can explicitly provide needed information to enhance the model's performance at inference. 
It also points out that instead of building models of ever increasing sizes, we can make a decent-size model output high-quality results by leveraging those training data that resemble the instance at hand. This can significantly reduce the computational cost while achieving a similar or better performance of a mega-sized model.

\section{Related Work}
\paragraph{Retrieval-based Methods} Even a pre-trained model as large as GPT-3~\cite{brown2020language} cannot remember everything, and it is important to leverage information retrieval to collect external knowledge to solve different NLP tasks. There are two types of representations for retriever: bag-of-word (BOW) based sparse representation~\cite{chen2017reading} and dense representation from neural networks~\cite{karpukhin2020dense}. 

For the sparse representation, as the method is based on BOW and usually rule-based score, such as BM25, is used for ranking, it can be easily adapted to a general large-scale search. This method has also been widely explored to solve open domain question answering~\cite{chen2017reading,wang2018r,lin2018denoising} and Machine Translation~\cite{gu2018search}. 

Dense representation based retrieval (DPR)~\cite{karpukhin2020dense} is the most widely explored area in recent years. Dense representations come from encoders, such as Transformer, trained with task-specific data. And these methods can achieve better recall performance than sparse representation on different tasks, such as open domain question answering~\cite{karpukhin2020dense,guu2020realm,yu2021kg}, knowledge-grounded generation~\cite{zhang2021joint}, and machine translation~\cite{cai2021neural}. One drawback of DPR is that it cannot process longer documents, usually less than 128 tokens~\cite{karpukhin2020dense}. Another drawback is that it needs parallel data for model training on specific tasks.

Considering the generalization and efficiency of sparse representation, in this paper, we use BM25 score~\cite{robertson2009probabilistic,schutze2008introduction} to retrieve from the training data, and our method is more flexible with no requirement of parallel data for model training.
Compared to non-parametric systems guided by search engine~\cite{gu2018search,khandelwal2019generalization}, our proposed method is based on supervised learning and is more general. 
\citet{lewis2021paq} is related to our work by retrieving related questions from pre-built large-scale question-answer pairs. However, our method doesn't need addition data augmentation method, and we have successfully applied \modelname to a wide range of downstream tasks, including summarization, question answering, machine translation and language modeling.

\paragraph{Prompt Engineering}
With the success of large-scale language models~\cite{brown2020language} on fewshot learning, prompt engineering comes to be a popular research direction. The idea is to prepend several labeled instances to the input sequence and then conduct the classification or generation.
\citet{liu2021makes} proposes to prepend the most related  labeled data as prompt to help fewshot inference.
\citet{li2021prefix} optimizes the prompt in continuous space. Motivated by these works where a good labeled prompt can help fewshot learning, we also prepend/append the most similar labeled training data for all the data in training, validation, and test set. However, different from prompt learning, we focus on supervised learning settings.

\section{Model}
\begin{figure*}[!ht]
  \centering
  \includegraphics[width=\linewidth]{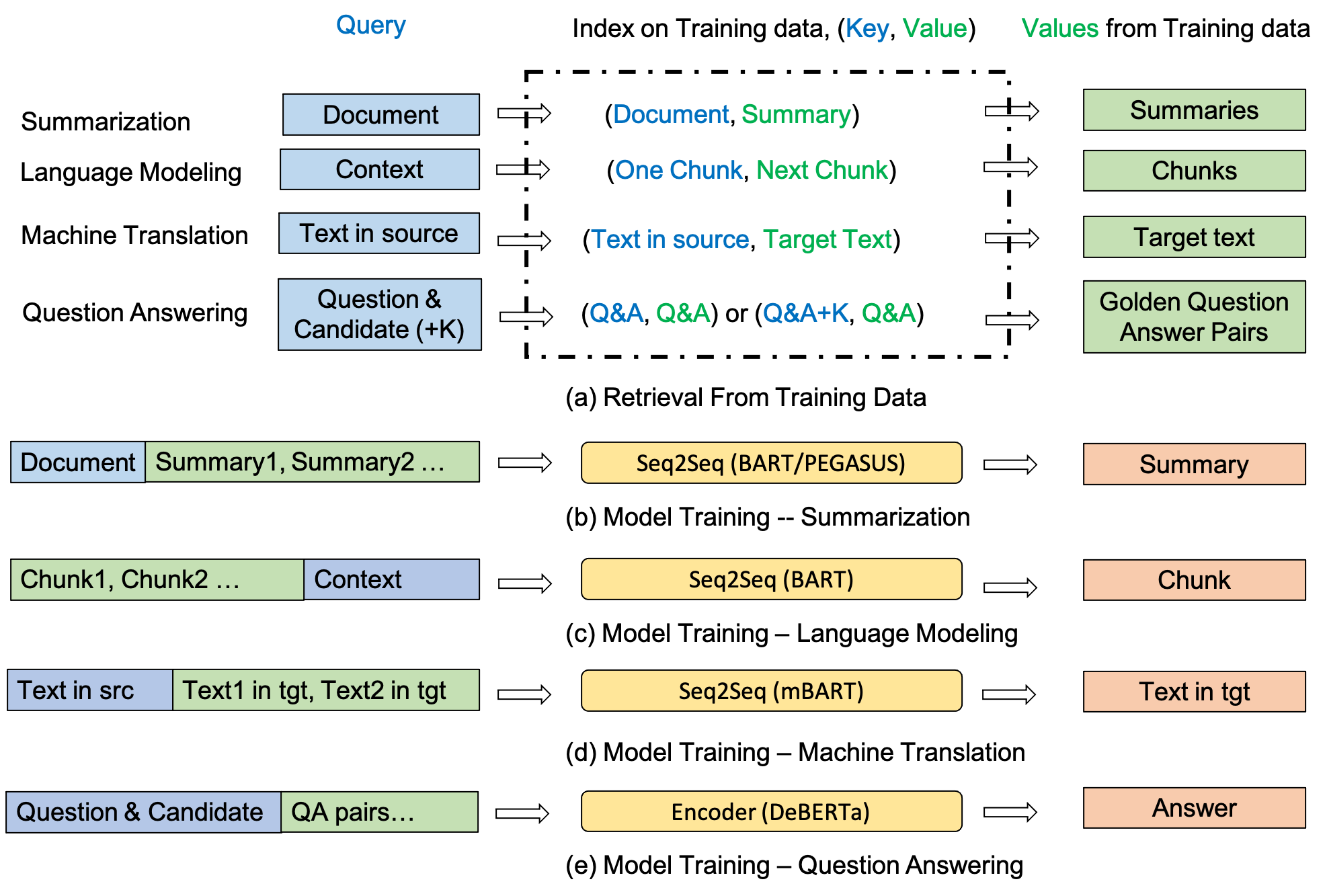}
  \caption{Model training with retrieval from the training data (\modelname). (a) Index on the training data and data retrieval for 4 different tasks. Box in blue is the query or the input sequence to encode. Box in green is the retrieved text. (b-e) Leveraging retrieved data for model training with different structures. For language modeling, we prepend the retrieved data to the query data, and append the retrieved data to the query for all the other tasks. After concatenation, we will directly feed them into Transformers, either Seq2Seq or Encoder-only frameworks, for text generation and answering selection. As we focus on the question answering tasks requiring commonsense reasoning, we have another version of index integrating knowledge graph for more precise retrieval. K: external knowledge from ConceptNet and Wiktionary, src: source language, tgt: target language.}
  \label{fig:model}
\end{figure*}
In this section, we will introduce the details of our proposed method. Briefly, given the input, we first retrieve the most matched instances with labels from the training data. We then concatenate them with the input sequence to feed into the model for generating the output. An overview of the whole method is shown in Figure~\ref{fig:model}.
\subsection{Retrieval-based Methods}
A retrieval-based method collects information most similar to the input from a corpus and then combines it with the input to feed into the NLP model. Suppose we index the corpus into a list of key-value pairs, i.e. $\mathcal{C}=\{(k_i, v_i)\}$. Then, given the input $x$, the retrieval engine $\mathcal{E}$ matches it with all keys and returns the top $K$ most similar keys to the query together with their values:
\begin{equation}
    \{(k_{i_1}, v_{i_1}), ..., (k_{i_K}, v_{i_K})\} = \mathcal{E}(x|\mathcal{C})
\end{equation}
In this work, we build the retrieval engine based on the widely used BM25 score \cite{schutze2008introduction}. We choose BM25 over dense representation mainly for its faster speed.

Then, these retrieved results are combined with the input $x$ to feed into the NLP model $\mathcal{M}$ to generate the output $O$:
\begin{equation}
    O=\mathcal{M}(f(x, \{(k_{i_1}, v_{i_1}), ..., (k_{i_K}, v_{i_K})\})
\end{equation}

Here, the combination function $f$ can be concatenation, e.g. $f(x, \{(k_{i_1}, v_{i_1}), ..., (k_{i_K}, v_{i_K})\}) = [x; v_{i_1}; ...; v_{i_K}]$. As data in different tasks is organized in different formats with varying lengths, we will introduce how we define different combination functions $f$ for various tasks in the follows.

\subsection{Retrieval from Training Data (\modelname)}
As retrieval from a large corpus is computationally costly, we propose to retrieve from the labeled training data. In other words, we directly adopt the training data $\mathcal{T}=\{(x_1, y_1), ..., (x_N, y_N)\}$ as the indexed corpus $\mathcal{C}$, where $x_i$ is the input and $y_i$ is the ground-truth label. 

Given an input $x$, the top $K$ retrieved training instances with labels are combined with $x$ as input to the model $\mathcal{M}$, i.e., $\mathcal{M}(f(x, \{(x_{i_1}, y_{i_1}), ..., (x_{i_K}, y_{i_K})\}$. Both training and inference take this retrieve-combine-generate scheme. Note that during training, as the input $x$ is already indexed, we filter it from the retrieval results to avoid data leakage.

Now, we introduce how we define the keys, values, and the combination function for different NLP tasks.

\textbf{Summarization} is to generate a summary for a given document. 
We first build an index for the document-summary pairs in the training data, where a document is the key and its summary is the value. Given a document $x$, we search for the most similar documents in the index. As documents are usually quite long, the combination function only keeps the values (summaries), i.e.,
$f_{summ}(x, \{(x_{i_1}, y_{i_1}), ..., (x_{i_K}, y_{i_K})\})=[x; y_{i_1}; ...; y_{i_K}]$.

\textbf{Language Modeling} (LM) generates the probability of a given sequence of words.
Typically, a Left-to-Right language model~\cite{dong2019unified} is trained on chunked sequences with an attention mask. 
In this paper, we use Seq2Seq based approach, i.e., given a context chunk, we predict the next chunk of text. 

In detail, we first chunk all the text in the training data. The IR index is built with one chunk $C_i$ as the key $x_i$ and its next chunk $C_{i+1}$ as the value $y_i$. Given a chunk $x$, we look for the most similar keys in the index and prepend their corresponding next chunks to $x$, i,e., $f_{LM}(x, \{(x_{i_1}, y_{i_1}), ..., (x_{i_K}, y_{i_K})\})=[y_{i_1}; ...; y_{i_K}; x]$.

\textbf{Machine Translation} is to translate text from the source language $\mathcal{S}$ to the target language $\mathcal{T}$. We define the key to be the sentence in $\mathcal{S}$ and the value to be its translation in $\mathcal{T}$. To keep the sequence short and speed up the training process, we only concatenate the retrieved text in target language:
$f_{MT}(x, \{(x_{i_1}, y_{i_1}), ..., (x_{i_K}, y_{i_K})\})=[x;  y_{i_1}; ...;  y_{i_K}]$.

\textbf{Question Answering} 
We mainly consider multiple-choice question answering, where commonsense knowledge is also required to reach the correct answer.
For each question $x_i$, there is a correct choice $y_i$ and several distractive candidate choices.
We index the concatenation of the question and the corresponding  ground-truth choice.
For a new question $x$, the model is given several choices $c_1, ..., c_M$. We concatenate $x$ with each choice $c_i$ as the query and retrieve related training instances:
$\{(x_{i_1}, y_{i_1}), ..., (x_{i_K}, y_{i_K})\} = \mathcal{E}(x; c_i|\mathcal{C})$. The combination function $f$ concatenates both retrieved question and answers with the input:
$f_{QA}((x, c_i), \{(x_{i_1}, y_{i_1}), ..., (x_{i_K}, y_{i_K})\})=[x; c_i; x_{i_1}; y_{i_1}; ...; x_{i_K}; y_{i_K}]$. Then, the model predicts a score representing how likely $c_i$ is the correct choice to $x$.

As the task requires commonsense knowledge, we build another version of index integrating commonsense knowledge. We follow the strategy from \cite{xu2021fusing} and extract the knowledge from ConceptNet \cite{speer2017conceptnet} and Wiktionary\footnote{\url{https://www.wiktionary.org/}} for the concepts in the question and choices.
% The knowledge, denoted by $\mathcal{K}$, includes..... \shuo{TODO}
For each question $x$ and choice $c$, we use string match to find corresponding entities in ConceptNet: $E^{(x)} = \{e^{(x)}_1,...,e^{(x)}_{n_x}\}$ appears in the question, and $E^{(c)} = \{e^{(c)}_1,...,e^{(c)}_{n_c}\}$ appears in the answer. To find the most relevant concept, we choose the concept with maximum length as the question and answer concept.
%: $e^{(x)} = \argmax_{e\in e^{(x)}} l(e), e^{(c)} = \argmax_{e\in e^{(c)}} l(e)$. 
We find the definition of the chosen concepts from Wiktionary. To find relations in ConceptNet, we find edges that connects question and answer concepts: $R = \{(e_1, r, e_2) | e_1\in E^{(x)}, e_2\in E^{(c)}, (e_1, e_2)\in \mathcal{KG}\}$. Here $\mathcal{KG}$ is ConceptNet and $r$ is a relation (e.g., \texttt{AtLocation}). 
We concatenate the Wiktionary definitions and ConceptNet relations $R$ to form the knowledge, $\mathcal{K}$, for a question.
The knowledge $\mathcal{K}$ is included both in the query and index. Thus, the retrieval process becomes:
$\{(x_{i_1}, c_{i_1}, \mathcal{K}_{i_1}), ..., (x_{i_K}, y_{i_K}, \mathcal{K}_{i_K})\} = \mathcal{E}(x; c_i; \mathcal{K}|\mathcal{C})$. The combination function $f$ concatenates retrieved questions and answers with the input:
$f_{QAK}((x, c_i), \mathcal{E}(x; c_i; \mathcal{K}|\mathcal{C}))=[x; c_i; x_{i_1}; y_{i_1}; ...; x_{i_K}; y_{i_K}]$.

\subsection{Model Training and Inference}
After concatenating the input with the retrieved data from the training corpus, we  feed the new sequence into the Seq2Seq framework for generation tasks and the encoder-only framework for question answering tasks. During training, as it will also  retrieve the exact golden label, we filter it directly. During inference, we will not filter any retrieved information, as all the retrieve data only come from training set. 
\section{Experiment}

\begin{table}[]

\setlength{\tabcolsep}{0.9mm}
\begin{tabular}{llccc}
\toprule
\multicolumn{1}{l}{Task}                                                       & Dataset       & Train & Dev & Test \\ \midrule
\multirow{5}{*}{\begin{tabular}[c]{@{}c@{}}Summar-\\ ization\end{tabular}}  & Multi-News    & 45k    & 5.6k & 5.6k  \\   
                                                                               & WikiHow       & 168k   & 6k   & 6k    \\
                                                                               &  XSum          & 204k   & 11k  & 11k   \\
                                                                               & NEWSROOM      & 993k   & 108k & 108k  \\ 
                                                                               & BigPatent     & 1,207k  & 67k  & 67k   \\\midrule
\multirow{2}{*}{\begin{tabular}[c]{@{}c@{}}Language \\ Modeling\end{tabular}}  & WikiText2     & 32k    & 3.3k & 3.8k  \\
& WikiText103   & 801k   & 1.7k & 1.9k  \\
                                                                                \midrule
\multirow{2}{*}{\begin{tabular}[c]{@{}c@{}}Machine\\ Translation\end{tabular}} & WMT16 (en-tr) & 205k   & 1k   & 3k    \\
                                                                               & WMT16 (en-de) & 4,548k  & 2.2k & 3k    \\ \midrule
\multirow{3}{*}{\begin{tabular}[c]{@{}c@{}}Question \\ Answering\end{tabular}} & CSQA &   9.7k    & 1.2k    &  1.1k    \\
                                                                               & PIQA          &  16k     &  1.8k   &  3.4k   \\
                                                                               & aNLI          & 170k       &  1.5k   &  3.0k    \\ \bottomrule
\end{tabular}
\caption{Statics of the evaluation datasets. The table shows the number of data in training, dev, and test sets. As we treat the language model as a Seq2Seq problem, the number here is the chunked sequences, each of which contains 64 words for WikiText2 and 128 words for WikiText103. }
\label{tbl:stats}
\end{table}

\begin{table*}[]

\setlength{\tabcolsep}{1.2mm}
\begin{tabular}{lccccccccccccccc}
\toprule
         & \multicolumn{3}{c}{BigPatent}               & \multicolumn{3}{c}{XSum}                      & \multicolumn{3}{c}{WikiHow}                   & \multicolumn{3}{c}{Multi-News}                 & \multicolumn{3}{c}{NEWSROOM}                  \\
         & R-1           & R-2           & R-L           & R-1           & R-2           & R-L           & R-1           & R-2           & R-L           & R-1           & R-2           & R-L           & R-1           & R-2           & R-L           \\
         \midrule
Earlier SOTA & 37.5&10.6&22.7 & 45.1&22.2&37.2& 28.5&9.2&26.5 & 43.4&14.8&17.4 &  39.9&28.3&36.8\\
PEGASUS     & 53.6          & 33.2          & 42.3          & 47.2          & 24.6          & 39.3          & 43.1          & 19.7          & 34.8          & \textbf{47.5} & \textbf{18.7} & \textbf{24.9} & \textbf{45.2} & \textbf{33.5} & \textbf{41.3} \\ 
\midrule \midrule
PEGASUS     & 38.4          & 13.5          & 26.3          & 46.6          & 23.9          & 38.6          & 35.9          & 15.3          & 30.3          & 43.1          & 15.4          & 22.6          & 41.7          &  30.7          & 37.8          \\
\modelname (PG) & 44.6          & 21.5          & 33.0          & \textbf{48.2} & \textbf{26.0} & \textbf{40.2} & 36.8          & 16.7          & 31.0          & 45.0          & 17.1          & 23.8          & 41.4          & 30.5          & 37.5          \\
BART-base     & 44.2          & 16.9          & 28.4          & 41.0          & 18.2          & 33.3          & 43.3          & 18.1          & 33.9          & 44.8          & 16.4          & 23.3          & 41.3          & 29.1          & 37.5          \\
\modelname (B) & 59.5          & 42.6          & 50.6          & 43.2          & 21.0          & 35.5          & 44.2 & 19.4          & 34.9          & 45.1          & 16.9          & 23.6          & 41.2          & 29.0          & 37.5          \\
BART-large     & 44.9          & 17.5          & 28.9          & 44.7          & 21.6          & 36.5          & 43.4          & 19.0          & 34.9          & 44.1          & 16.6          & 22.7          & 41.6          & 29.4          & 38.0          \\
\modelname (L) & \textbf{60.7} & \textbf{43.3} & \textbf{51.3} & {\ul 46.5 }         & {\ul 24.1}          & {\ul 38.6}          & \textbf{44.2}          & \textbf{20.4} & \textbf{35.8} & {\ul 46.9}    & {\ul 17.7}    & {\ul 24.0}    & {\ul 42.5}    &  {\ul 30.2}    & {\ul 38.7}   \\
\bottomrule
\end{tabular}
\caption{Summarization results. In the top section, we report the results from PEGASUS~\cite{zhang2020pegasus} paper. In the bottom, we reproduce three strong baselines with PEGASUS and BART~\cite{lewis2019bart}, and show our \modelname initialized by the same pre-trained models for fair comparison. The bolded numbers show the SOTA performance and the underlined numbers show the best performance with BART initialization. PEGASUS: PEGASUS-large, B: BART-base, L: BART-large, R-1: Rouge-1, R-2: Rouge-2, R-L: Rouge-L}
\label{tbl:summarization}
\end{table*}
In this section, we will introduce more details about experiments and the corresponding analysis. 
\subsection{Dataset}
We evaluate \modelname on 4 different tasks with 12 datasets as shown in Table~\ref{tbl:stats}.
\paragraph{Summarization} We evaluate our method on 5 summarization datasets: 1) \textbf{XSum}~\cite{narayan2018don}, extreme summarization, is a task of one sentence summarization on one document. The document comes from  British Broadcasting Corporation (BBC) online articles. 2) \textbf{NEWSROOM}~\cite{grusky2018newsroom} is a summarization
dataset on a larger scale and the articles with human-written summaries come from 38 major news publications. 
 3) \textbf{Multi-News}~\cite{fabbri2019multi} is a task of multi-document summarization on news articles from the site newser.com. 
 4) \textbf{BigPatent}~\cite{sharma2019bigpatent}  is constructed on U.S. patent documents along with human written abstracts. The documents cover broader areas in 9 different categories.
Another domain, 5) \textbf{WikiHow}~\cite{koupaee2018wikihow} is to summarize the steps
of ``How to" solve a problem. The dataset consists of more diverse style articles written by ordinary people. Besides the above datasets, we also introduce CNN/Dailymail~\cite{nallapati2016abstractive} and 160G BART pretraining corpus~\cite{lewis2019bart} from BOOKCORPUS, CC-NEWS, OPENWEBTEXT, and STORIES,  to scale up the training corpus.

\paragraph{Language Modeling}
As our model is initialized by a pre-trained model, we select two language modeling datasets, the corpus of which is not used for model pre-training. The text of both datasets, \textbf{WikiText103}~\cite{merity2016pointer} and \textbf{WikiText2}~\cite{merity2016pointer}, are extracted from Wikipedia. As the dataset's text is at a document level, the tasks focus on testing the model's ability to remember longer sequences.
 
\paragraph{Machine Translation}
We evaluate our method on the translation of English-German and English-Turkish in both directions from WMT16~\cite{bojar2016findings}. 

\paragraph{Question Answering}
We have 3 question answering datasets to evaluate our method: 1) \textbf{CommonsenseQA} (CSQA, \citeauthor{talmor2018commonsenseqa}, \citeyear{talmor2018commonsenseqa})
% ~\cite{talmor2018commonsenseqa} 
is a dataset for commonsense multi-choice
question answering. The questions are generated based on commonsense knowledge base, ConceptNet. 2) \textbf{Physical IQA} (PIQA, \citeauthor{bisk2020piqa}, \citeyear{bisk2020piqa}) 
% \textbf{PIQA}~\cite{bisk2020piqa} 
is to answer questions requiring physical commonsense reasoning. 3)
\textbf{Abductive NLI} (aNLI, \citeauthor{bhagavatula2019abductive}, \citeyear{bhagavatula2019abductive}) 
% \textbf{ANLI}~\cite{bhagavatula2019abductive}, Abductive NLI, 
is a multiple-choice question answering task for choosing the more likely explanation. All these tasks are challenging by requiring commonsense knowledge to reach the correct answer.

\begin{comment}
\begin{table}[]
\setlength{\tabcolsep}{1.3mm}
\begin{tabular}{lccccc}
\toprule
                & \multicolumn{3}{c}{XSUM} & CSQA  & CSQA* \\
                & R-1    & R-2    & R-L    & Acc   & Acc     \\ \midrule
Baseline       & 44.6       &  21.6      &  36.9   &\multicolumn{2}{c}{88.56}  \\
\modelname (STask)  & 46.5   & 24.1   & 38.6   & 90.4 & 89.8   \\
\modelname (MTask-1) & 47.5  & 25.2  & 39.5  & 90.1 & 90.5   \\
\modelname (MTask-2) & 47.5  & 24.9  & 39.4  & 88.0 & 90.8  \\ \bottomrule
\end{tabular}
\caption{Mult-Task results. For CSQA, MTask-RFT-1 retrieves data from CSQA, OBQA and RiddleSense. MTask-RFT-2 retrieves from a collection of 16 commonsense reasoning datasets.}
\end{table}
\end{comment}

\begin{table}[]
\centering
\setlength{\tabcolsep}{1.7mm}
\begin{tabular}{lccc}
\toprule
                & \multicolumn{3}{c}{XSum}  \\
                & R-1    & R-2    & R-L         \\ \midrule
BART (XSum)       & 44.7       &  21.6      &  36.5   \\
BART (XSum+CNN)       & 44.6       &  21.6      &  36.9   \\
\modelname (XSum)  & 46.5   & 24.1   & 38.6     \\
\modelname (XSum+CNN) & 47.5  & \textbf{25.2}  & \textbf{39.5}     \\ 
\modelname (XSum+NR) & 47.5  & 24.9  & 39.4    \\ 
\modelname (XSum+160G) & \textbf{47.7}&	25.1&	\textbf{39.5} \\ \bottomrule
\end{tabular}
\caption{Evaluation on XSum test set with training data scale up. BART is jointly trained with datasets in bracket. \modelname is trained with  XSum document-summary pairs, but the index is built on the datasets in bracket. CNN: CNN/Dailymail dataset, NR: NEWSROOM dataset, 160G: BART pre-training corpus.  }
\label{tbl:multitask}
\end{table}

\subsection{\modelname Details}

For the task of summarization, instead of directly retrieving the most relevant summary~\cite{an2021retrievalsum}, we find the most relevant documents by BM25 score and then leverage the corresponding summaries.  
Compared to the dense passage retrieval based method, our method can handle the long document retrieval and does not need to train. Moreover, \modelname is easier to scale up. We also consider joint training baseline on Summarization tasks. Our setting is to test how other datasets can help improve XSum. For REINA, we build index on summarization datasets from different sources. During model training, we will only train models with the XSum dataset along with retrieved data appended to the documents.

\begin{table}[t]
\centering
\begin{tabular}{lccc}
\toprule
            & CSQA & aNLI & PIQA \\ \midrule
\multicolumn{4}{l}{Dev Set results}\\
DeBERTa    & 84.0 & 88.8    &  85.6    \\
\modelname (w/o K)  & \textbf{88.8} & 88.6    &  85.5    \\ 
\modelname (w/ K) & 86.8 & \textbf{89.6}    &  \textbf{86.9 }   \\ 
\midrule
\multicolumn{4}{l}{Test Set results}\\
CALM  & 71.8 & 82.4  & 76.9  \\
UNICORN   & 79.3 & 87.3 & \textbf{90.1}     \\
DEKCOR      & 83.3 &  -    &  -    \\ 
DeBERTa     & - & 86.8    &  85.1    \\
\modelname & \textbf{84.6} & \textbf{88.0}   &  {\ul 85.4}   \\ \bottomrule
\end{tabular}
\caption{Question answering results. CALM~\cite{zhou2021pretraining} is continue-pretrained from RoBERTa-large model. UNICORN~\cite{Lourie2021UNICORNOR} and DEKCOR~\cite{xu2021fusing} use the T5-11B model. Our DeBERTa baseline is close to DEKCOR but with different pretrained initializations. \modelname is also based on DeBERTa. We first evaluate \modelname on dev set to verify whether integrating external knowledge in \modelname can lead to better performance. And then submit the best one for hidden test set evaluation. We achieve leaderboard No.1 on CommonsenseQA.  K: external knowledge from ConceptNet and Wiktionary.  
% CSQA: CommonsenseQA
}
\label{tbl:csqa}
\end{table}
For language modeling task, instead of working on word-level retrieval by KNN~\cite{khandelwal2019generalization}, we chunk all the training data.  
During training, besides the retrieved chunks, we will also include the context of the query chunk to generate next chunk. Compared to KNN-LM~\cite{khandelwal2019generalization}, \modelname only needs retrieval once per chunk which is much more efficient. 
\begin{table}[ht]
\centering
\begin{tabular}{lcc}
\toprule
               & WikiText103    & WikiText2      \\
               \midrule
Transformer-XL &  18.30 & - \\               
kNN-LM         & 15.79 &  -   \\  
GPT-2          & 17.48          & 18.34          \\
\midrule \midrule
BART-Base      & 15.88          & 20.41          \\
\modelname (B)  & 14.76          & 20.78          \\
BART-Large     & 12.10          & \textbf{15.11} \\
\modelname (L) & \textbf{11.36} & 15.62         \\
\bottomrule
\end{tabular}
\caption{Language modeling results. The evaluation metric is perplexity (PPL). The top part of the table comes from the original papers, Transformer-XL~\cite{dai2019transformer}, kNN-LM~\cite{khandelwal2019generalization}, GPT-2~\cite{radford2019language}. The bottom part is our implementation with fair comparison.  B: BART-base, L: BART-large}
\label{tbl:lm}
\end{table}

For multi-choice question answering, we build two types of indexes with or without external knowledge from ConceptNet and Wiktionary. For the query, the concatenation of question and one candidate answer, we also have two versions, with or without knowledge. After adding knowledge, there would be more word overlaps when key concept words between questions are matched.
The retrieved information will be treated as either a prompt or additional knowledge to encode together and then predicts the answer probability of each candidate.

\subsection{Optimization Details}
Our information retrieval is based on Lucene Index~\footnote{https://lucene.apache.org/pylucene/}. Our model training is based on Transformers library~\footnote{https://github.com/huggingface/transformers}. All our experiments are based on 8-GPU machines.

For summarization tasks, we initialized the model with three types of pre-trained models, PEGASUS-large~\cite{zhang2020pegasus}, BART-base, and BART-large~\cite{lewis2019bart}. Optimization is based on AdamW~\cite{loshchilov2017decoupled}. We tune the hyper-parameters from learning rate \{2e-05, 5e-05, 7e-05\}, and set dropout 0.1, batch size 32. For both baseline and our method, we set the maximal length of the input sequence to be 1024. We use the original document to generate summary in baselines. For REINA, we set the maximal length of the original document 600 and then append the top-5 retrieved summaries from training data.  

For language modeling tasks, we initialized the model with BART-base and BART-large. 
We set the number of words in each chunk to 128 for WikiText103 and 64 for WikiText2. For each chunk generation, we set the context length of baseline methods 1024. For our method, we set the context 512 and prepend the retrieved text. The maximal length of the concatenated sequence is 1024. We use optimizer  Adam~\cite{kingma2014adam} with learning rate 5e-05, dropout 0.1, batch size 32.

For machine translation tasks, we initialized the model with mBART-large~\cite{liu2020multilingual}. We follow the hyper-parameter setting from the original paper with Adam optimizer, dropout 0.3, label smoothing 0.2,  warm-up steps 2500,  maximum learning rate 3e-05, and training updates 40K in total.

For question answering datasets, our method is based on DeBERTa~\cite{he2020deberta} with 1.5B parameters. We use optimizer AdamW~\cite{loshchilov2017decoupled} with learning rate 3e-06, batch size 8. As the datasets requiring commonsense reasoning, we also leverage knowledge bases, ConceptNet and Wiktionary, in\modelname.
\begin{table}[]
\centering
\begin{tabular}{lcccc}
\toprule
          & \multicolumn{4}{c}{WMT16}                             \\
          & en2tr         & tr2en         & en2de         & de2en \\ \midrule
XLM & - & - & 26.4 & 34.3 \\ 
mBART     & 18.4          & 23.1          & 32.6          & 37.0  \\
\modelname  & \textbf{18.8} & \textbf{23.6} & \textbf{32.9} & 37.0  \\ \bottomrule
\end{tabular}
\caption{Machine translation on WMT16. We compare with baselines XLM~\cite{lample2019cross} and mBART~\cite{liu2020multilingual}. \modelname is initialized by mBART for fair comparison. The evaluation metric is based on SacreBLEU. Source and target languages are concatenated by ``2". tr: Turkish, de: German, en: English. }
\label{tbl:mt}
\end{table}

\subsection{Experiment Results}
Our experiment results on the summarization tasks are shown in Table~\ref{tbl:summarization}. Our evaluation metric is based on Rouge-1/2/L scores, same as PEGASUS~\cite{zhang2020pegasus}. We have a broad experiment on 5 datasets, ranging from single document summarization (XSum) to multi-document summarization (Multi-News), from news domain to wiki knowledge (WikiHow) and patent (BigPatent) domains. We re-run all of our baseline methods. Based on the experiment results, we find that \modelname can significantly boost the baselines initialized with different pre-trained models, such as PEGASUS, BART-base, and BART-large, on all 5 datasets. Besides, our method with BART-large can achieve state-of-the-art performance on XSum and BigPatent datasets. Moreover, we find \modelname can help base models beat larger models. For example, \modelname (BART-base) is better than both PEGASUS-LARGE and BART-large on  BigPatent and WikiHow datasets. 

\begin{table*}[]
\centering
\small
\begin{tabular}{lp{13.5cm}}
\toprule
 Document & No international side has toured Bangladesh since 20 people were killed in a siege at a cafe in Dhaka in July.The England and Wales Cricket Board said in August that tour would go ahead following a security review ... \\
 Summary & England one-day captain Eoin Morgan and opening batsman Alex Hales have opted out of October's tour of Bangladesh because of security concerns.  \\
               \modelname 1   & England one-day captain Eoin Morgan says he will never again go on a tour where security concerns may affect his game. \\
 \modelname 2 & Eoin Morgan and Alex Hales remain "very much part of the group" despite not touring Bangladesh, says stand-in England one-day captain Jos Buttler.\\      
 \midrule \midrule
 Question &  Brawn opened the curtains so that the sun could do what?  \\
 Answer & \modelname chooses: \textbf{warm room},\; Baseline chooses: \emph{shine brightly}  \\
 \modelname  1   & What effect did the sun have on the residents inside? warm house.  \\
 \modelname 2  &  James installed his new curtains to keep the light from shinning on his television.  Where is James probably hanging his curtains? house. \\
 \bottomrule
\end{tabular}
\caption{Examples from dev sets and the corresponding labeled data retrieved from training set. The top case comes from a summarization task, XSum. The bottom case comes from a question answering task, CommonsenseQA. For summarization tasks, we will only append the document with the retrieved summaries. For CommonsenseQA, we will append the golden QA pairs to the question. The golden answer is ``warm room". \modelname 1/2 refers to different retrieved data.
}
\end{table*}
We also evaluate the ability of \modelname on learning from more related datasets. Our experiment results are shown in Table~\ref{tbl:multitask}. The evaluation is conducted on XSum test set and we use three related data sources from CNN/Dailymail, NEWSROOM, and a 160G raw-text corpus\footnote{For the 160G data, we treat the first sentence as summary and the rest as document.}. Based on the experiments, we can see that simply training the model on merged dataset (XSum + other sources) doesn't lead to any gains. However, after adding one additional data source to build index and applying REINA, there's 1\% improvement in Rouge scores\footnote{In our experiments, we follow \citet{xu2021dissecting} by ignoring the retrieved data if there are over three 7-gram overlap between retrieved summary and golden summary.}. Overall, our \modelname can effectively leverage the most relevant data from additional datasets while being trained only on the target task.

For question answering tasks, our results are shown in Table~\ref{tbl:csqa}. We test \modelname on three datasets, where commonsense knowledge is usually required to answer the question. Thus we first verify whether we need external knowledge during the retrieval. According to the experiments, we find that directly retrieving the labeled data without knowledge works best for CommonsenseQA dataset, but involving knowledge can help on aNLI and PIQA datasets. And \modelname can significantly improve our baselines with DeBERTa on all the datasets. Moreover, after submitting our best results to the corresponding leaderboards, \modelname  achieves state of the art on CommonsenseQA dataset (Leaderboard No.1) and beat strong baselines on aNLI and PIQA datasets.

Our evaluation of language modeling is shown in Table~\ref{tbl:lm}. Our method can achieve significant improvement on WikiText103 dataset over both BART-base and BART-large baselines. However, it cannot lead to better performance on WikiText2. One reason may be that WikiText2 is a much smaller dataset, and it's hard for \modelname to retrieve the most related text. Besides, we also find Seq2Seq model can be a very strong baseline which means we can leverage more pre-trained models such as PEGASUS, T5~\cite{raffel2019exploring},  and BART, for language modeling in future work. And Seq2Seq frame would be more flexible to integrate external knowledge to  boost performance further.

For machine translation, we make use of the datasets from WMT16. We select one low-resource language, Turkish-English, and one rich-resource, German-English, for \modelname evaluation, as shown in Table~\ref{tbl:mt}. We re-implement mBART baseline for translation in both directions. To make a fair comparison, \modelname is also based on mBART. We can find that \modelname can further boost performance under three settings, translating English to Turkish, Turkish to English, and English to German. 

\subsection{Further Analysis}
We show a case study on the data retrieved by \modelname. We list two cases from XSum and CommonsenseQA dev sets. From the case on summarization task, we can find that the first retrieved summary from training set, \modelname 1, shows the same point of ``security concerns" as the golden summary. And the other case on multi-choice question answering, \modelname 1 suggests that the sun can warm up a place that shares the same commonsense knowledge to answer the question. After, although we cannot visualize how the neural encoders work by leveraging the retrieved data, we have shown that the data from \modelname have very strong correlation with the golden labels. 
\section{Conclusion}
In this paper, we propose a simple and effective method to fully make use training dataset. Our proposed method is  general  and can be easily integrated into different models on different tasks. We prove that \modelname can effectively improve baseline performance on 11 datasets covering summarization, language modeling, machine translation, and question answering tasks.

\bibliography{anthology,custom}

\begin{thebibliography}{46}
\expandafter\ifx\csname natexlab\endcsname\relax\def\natexlab#1{#1}\fi

\bibitem[{An et~al.(2021)An, Zhong, Geng, Yang, and Qiu}]{an2021retrievalsum}
Chenxin An, Ming Zhong, Zhichao Geng, Jianqiang Yang, and Xipeng Qiu. 2021.
\newblock Retrievalsum: A retrieval enhanced framework for abstractive
  summarization.
\newblock \emph{arXiv preprint arXiv:2109.07943}.

\bibitem[{Bhagavatula et~al.(2020)Bhagavatula, Bras, Malaviya, Sakaguchi,
  Holtzman, Rashkin, Downey, Yih, and Choi}]{bhagavatula2019abductive}
Chandra Bhagavatula, Ronan~Le Bras, Chaitanya Malaviya, Keisuke Sakaguchi, Ari
  Holtzman, Hannah Rashkin, Doug Downey, Scott Wen-tau Yih, and Yejin Choi.
  2020.
\newblock Abductive commonsense reasoning.
\newblock \emph{International Conference on Learning Representations (ICLR)}.

\bibitem[{Bisk et~al.(2020)Bisk, Zellers, Gao, Choi et~al.}]{bisk2020piqa}
Yonatan Bisk, Rowan Zellers, Jianfeng Gao, Yejin Choi, et~al. 2020.
\newblock Piqa: Reasoning about physical commonsense in natural language.
\newblock In \emph{AAAI Conference on Artificial Intelligence (AAAI)}.

\bibitem[{Bojar et~al.(2016)Bojar, Chatterjee, Federmann, Graham, Haddow, Huck,
  Yepes, Koehn, Logacheva, Monz et~al.}]{bojar2016findings}
Ond{\v{r}}ej Bojar, Rajen Chatterjee, Christian Federmann, Yvette Graham, Barry
  Haddow, Matthias Huck, Antonio~Jimeno Yepes, Philipp Koehn, Varvara
  Logacheva, Christof Monz, et~al. 2016.
\newblock Findings of the 2016 conference on machine translation.
\newblock In \emph{First Conference on Machine Translation}.

\bibitem[{Brown et~al.(2020)Brown, Mann, Ryder, Subbiah, Kaplan, Dhariwal,
  Neelakantan, Shyam, Sastry, Askell et~al.}]{brown2020language}
Tom~B Brown, Benjamin Mann, Nick Ryder, Melanie Subbiah, Jared Kaplan, Prafulla
  Dhariwal, Arvind Neelakantan, Pranav Shyam, Girish Sastry, Amanda Askell,
  et~al. 2020.
\newblock Language models are few-shot learners.
\newblock \emph{Neural Information Processing Systems (NeurIPS)}.

\bibitem[{Cai et~al.(2021)Cai, Wang, Li, Lam, and Liu}]{cai2021neural}
Deng Cai, Yan Wang, Huayang Li, Wai Lam, and Lemao Liu. 2021.
\newblock Neural machine translation with monolingual translation memory.
\newblock \emph{Association for Computational Linguistics (ACL)}.

\bibitem[{Chen et~al.(2017)Chen, Fisch, Weston, and Bordes}]{chen2017reading}
Danqi Chen, Adam Fisch, Jason Weston, and Antoine Bordes. 2017.
\newblock Reading wikipedia to answer open-domain questions.
\newblock \emph{Association for Computational Linguistics (ACL)}.

\bibitem[{Dai et~al.(2019)Dai, Yang, Yang, Carbonell, Le, and
  Salakhutdinov}]{dai2019transformer}
Zihang Dai, Zhilin Yang, Yiming Yang, Jaime Carbonell, Quoc~V Le, and Ruslan
  Salakhutdinov. 2019.
\newblock Transformer-xl: Attentive language models beyond a fixed-length
  context.
\newblock \emph{Association for Computational Linguistics (ACL)}.

\bibitem[{Dong et~al.(2020)Dong, Yang, Wang, Wei, Liu, Wang, Gao, Zhou, and
  Hon}]{dong2019unified}
Li~Dong, Nan Yang, Wenhui Wang, Furu Wei, Xiaodong Liu, Yu~Wang, Jianfeng Gao,
  Ming Zhou, and Hsiao-Wuen Hon. 2020.
\newblock Unified language model pre-training for natural language
  understanding and generation.
\newblock \emph{International Conference on Machine Learning (ICML)}.

\bibitem[{Fabbri et~al.(2019)Fabbri, Li, She, Li, and Radev}]{fabbri2019multi}
Alexander~R Fabbri, Irene Li, Tianwei She, Suyi Li, and Dragomir~R Radev. 2019.
\newblock Multi-news: A large-scale multi-document summarization dataset and
  abstractive hierarchical model.
\newblock \emph{Association for Computational Linguistics (ACL)}.

\bibitem[{Grusky et~al.(2018)Grusky, Naaman, and Artzi}]{grusky2018newsroom}
Max Grusky, Mor Naaman, and Yoav Artzi. 2018.
\newblock Newsroom: A dataset of 1.3 million summaries with diverse extractive
  strategies.
\newblock \emph{North {A}merican Chapter of the Association for Computational
  Linguistics (NAACL)}.

\bibitem[{Gu et~al.(2018)Gu, Wang, Cho, and Li}]{gu2018search}
Jiatao Gu, Yong Wang, Kyunghyun Cho, and Victor~OK Li. 2018.
\newblock Search engine guided neural machine translation.
\newblock In \emph{AAAI Conference on Artificial Intelligence (AAAI)}.

\bibitem[{Guu et~al.(2018)Guu, Hashimoto, Oren, and Liang}]{guu2018generating}
Kelvin Guu, Tatsunori~B Hashimoto, Yonatan Oren, and Percy Liang. 2018.
\newblock Generating sentences by editing prototypes.
\newblock \emph{Transactions of the Association for Computational Linguistics
  (TACL)}.

\bibitem[{Guu et~al.(2020)Guu, Lee, Tung, Pasupat, and Chang}]{guu2020realm}
Kelvin Guu, Kenton Lee, Zora Tung, Panupong Pasupat, and Ming-Wei Chang. 2020.
\newblock Realm: Retrieval-augmented language model pre-training.
\newblock \emph{International Conference on Machine Learning (ICML)}.

\bibitem[{He et~al.(2021)He, Liu, Gao, and Chen}]{he2020deberta}
Pengcheng He, Xiaodong Liu, Jianfeng Gao, and Weizhu Chen. 2021.
\newblock Deberta: Decoding-enhanced bert with disentangled attention.
\newblock \emph{International Conference on Learning Representations (ICLR)}.

\bibitem[{Karpukhin et~al.(2020)Karpukhin, O{\u{g}}uz, Min, Lewis, Wu, Edunov,
  Chen, and Yih}]{karpukhin2020dense}
Vladimir Karpukhin, Barlas O{\u{g}}uz, Sewon Min, Patrick Lewis, Ledell Wu,
  Sergey Edunov, Danqi Chen, and Wen-tau Yih. 2020.
\newblock Dense passage retrieval for open-domain question answering.
\newblock \emph{Empirical Methods in Natural Language Processing (EMNLP)}.

\bibitem[{Khandelwal et~al.(2021)Khandelwal, Fan, Jurafsky, Zettlemoyer, and
  Lewis}]{khandelwal2020nearest}
Urvashi Khandelwal, Angela Fan, Dan Jurafsky, Luke Zettlemoyer, and Mike Lewis.
  2021.
\newblock Nearest neighbor machine translation.
\newblock \emph{International Conference on Learning Representations (ICLR)}.

\bibitem[{Khandelwal et~al.(2020)Khandelwal, Levy, Jurafsky, Zettlemoyer, and
  Lewis}]{khandelwal2019generalization}
Urvashi Khandelwal, Omer Levy, Dan Jurafsky, Luke Zettlemoyer, and Mike Lewis.
  2020.
\newblock Generalization through memorization: Nearest neighbor language
  models.
\newblock \emph{International Conference on Learning Representations (ICLR)}.

\bibitem[{Kingma and Ba(2015)}]{kingma2014adam}
Diederik~P Kingma and Jimmy Ba. 2015.
\newblock Adam: A method for stochastic optimization.
\newblock \emph{International Conference on Learning Representations (ICLR)}.

\bibitem[{Koupaee and Wang(2018)}]{koupaee2018wikihow}
Mahnaz Koupaee and William~Yang Wang. 2018.
\newblock Wikihow: A large scale text summarization dataset.
\newblock \emph{arXiv preprint arXiv:1810.09305}.

\bibitem[{Lample and Conneau(2019)}]{lample2019cross}
Guillaume Lample and Alexis Conneau. 2019.
\newblock Cross-lingual language model pretraining.
\newblock \emph{Neural Information Processing Systems (NeurIPS)}.

\bibitem[{Lewis et~al.(2020)Lewis, Liu, Goyal, Ghazvininejad, Mohamed, Levy,
  Stoyanov, and Zettlemoyer}]{lewis2019bart}
Mike Lewis, Yinhan Liu, Naman Goyal, Marjan Ghazvininejad, Abdelrahman Mohamed,
  Omer Levy, Ves Stoyanov, and Luke Zettlemoyer. 2020.
\newblock Bart: Denoising sequence-to-sequence pre-training for natural
  language generation, translation, and comprehension.
\newblock \emph{Association for Computational Linguistics (ACL)}.

\bibitem[{Lewis et~al.(2021)Lewis, Wu, Liu, Minervini, K{\"u}ttler, Piktus,
  Stenetorp, and Riedel}]{lewis2021paq}
Patrick Lewis, Yuxiang Wu, Linqing Liu, Pasquale Minervini, Heinrich
  K{\"u}ttler, Aleksandra Piktus, Pontus Stenetorp, and Sebastian Riedel. 2021.
\newblock Paq: 65 million probably-asked questions and what you can do with
  them.
\newblock \emph{Transactions of the Association for Computational Linguistics}.

\bibitem[{Li and Liang(2021)}]{li2021prefix}
Xiang~Lisa Li and Percy Liang. 2021.
\newblock Prefix-tuning: Optimizing continuous prompts for generation.
\newblock \emph{Association for Computational Linguistics (ACL)}.

\bibitem[{Lin et~al.(2018)Lin, Ji, Liu, and Sun}]{lin2018denoising}
Yankai Lin, Haozhe Ji, Zhiyuan Liu, and Maosong Sun. 2018.
\newblock Denoising distantly supervised open-domain question answering.
\newblock In \emph{Association for Computational Linguistics (ACL)}.

\bibitem[{Liu et~al.(2021)Liu, Shen, Zhang, Dolan, Carin, and
  Chen}]{liu2021makes}
Jiachang Liu, Dinghan Shen, Yizhe Zhang, Bill Dolan, Lawrence Carin, and Weizhu
  Chen. 2021.
\newblock What makes good in-context examples for gpt-$3 $?
\newblock \emph{arXiv preprint arXiv:2101.06804}.

\bibitem[{Liu et~al.(2020)Liu, Gu, Goyal, Li, Edunov, Ghazvininejad, Lewis, and
  Zettlemoyer}]{liu2020multilingual}
Yinhan Liu, Jiatao Gu, Naman Goyal, Xian Li, Sergey Edunov, Marjan
  Ghazvininejad, Mike Lewis, and Luke Zettlemoyer. 2020.
\newblock Multilingual denoising pre-training for neural machine translation.
\newblock \emph{Transactions of the Association for Computational Linguistics
  (TACL)}.

\bibitem[{Loshchilov and Hutter(2019)}]{loshchilov2017decoupled}
Ilya Loshchilov and Frank Hutter. 2019.
\newblock Decoupled weight decay regularization.
\newblock \emph{International Conference on Learning Representations (ICLR)}.

\bibitem[{Lourie et~al.(2021)Lourie, {Le Bras}, Bhagavatula, and
  Choi}]{Lourie2021UNICORNOR}
Nicholas Lourie, Ronan {Le Bras}, Chandra Bhagavatula, and Yejin Choi. 2021.
\newblock Unicorn on rainbow: A universal commonsense reasoning model on a new
  multitask benchmark.
\newblock \emph{AAAI Conference on Artificial Intelligence (AAAI)}.

\bibitem[{Merity et~al.(2017)Merity, Xiong, Bradbury, and
  Socher}]{merity2016pointer}
Stephen Merity, Caiming Xiong, James Bradbury, and Richard Socher. 2017.
\newblock Pointer sentinel mixture models.
\newblock \emph{International Conference on Learning Representations (ICLR)}.

\bibitem[{Nallapati et~al.(2016)Nallapati, Zhou, Gulcehre, Xiang
  et~al.}]{nallapati2016abstractive}
Ramesh Nallapati, Bowen Zhou, Caglar Gulcehre, Bing Xiang, et~al. 2016.
\newblock Abstractive text summarization using sequence-to-sequence rnns and
  beyond.
\newblock \emph{SIGNLL Conference on Computational Natural Language Learning
  (CoNLL)}.

\bibitem[{Narayan et~al.(2018)Narayan, Cohen, and Lapata}]{narayan2018don}
Shashi Narayan, Shay~B Cohen, and Mirella Lapata. 2018.
\newblock Don't give me the details, just the summary! topic-aware
  convolutional neural networks for extreme summarization.
\newblock \emph{Empirical Methods in Natural Language Processing (EMNLP)}.

\bibitem[{Radford et~al.(2019)Radford, Wu, Child, Luan, Amodei, Sutskever
  et~al.}]{radford2019language}
Alec Radford, Jeffrey Wu, Rewon Child, David Luan, Dario Amodei, Ilya
  Sutskever, et~al. 2019.
\newblock Language models are unsupervised multitask learners.
\newblock \emph{OpenAI blog}.

\bibitem[{Raffel et~al.(2030)Raffel, Shazeer, Roberts, Lee, Narang, Matena,
  Zhou, Li, and Liu}]{raffel2019exploring}
Colin Raffel, Noam Shazeer, Adam Roberts, Katherine Lee, Sharan Narang, Michael
  Matena, Yanqi Zhou, Wei Li, and Peter~J Liu. 2030.
\newblock Exploring the limits of transfer learning with a unified text-to-text
  transformer.
\newblock \emph{Journal of Machine Learning Research (JMLR)}.

\bibitem[{Robertson and Zaragoza(2009)}]{robertson2009probabilistic}
Stephen Robertson and Hugo Zaragoza. 2009.
\newblock \emph{The probabilistic relevance framework: BM25 and beyond}.
\newblock Now Publishers Inc.

\bibitem[{Sch{\"u}tze et~al.(2008)Sch{\"u}tze, Manning, and
  Raghavan}]{schutze2008introduction}
Hinrich Sch{\"u}tze, Christopher~D Manning, and Prabhakar Raghavan. 2008.
\newblock \emph{Introduction to information retrieval}.
\newblock Cambridge University Press Cambridge.

\bibitem[{Sharma et~al.(2019)Sharma, Li, and Wang}]{sharma2019bigpatent}
Eva Sharma, Chen Li, and Lu~Wang. 2019.
\newblock Bigpatent: A large-scale dataset for abstractive and coherent
  summarization.
\newblock \emph{Association for Computational Linguistics (ACL)}.

\bibitem[{Speer et~al.(2017)Speer, Chin, and Havasi}]{speer2017conceptnet}
Robyn Speer, Joshua Chin, and Catherine Havasi. 2017.
\newblock Conceptnet 5.5: An open multilingual graph of general knowledge.
\newblock In \emph{AAAI conference on artificial intelligence (AAAI)}.

\bibitem[{Talmor et~al.(2019)Talmor, Herzig, Lourie, and
  Berant}]{talmor2018commonsenseqa}
Alon Talmor, Jonathan Herzig, Nicholas Lourie, and Jonathan Berant. 2019.
\newblock Commonsenseqa: A question answering challenge targeting commonsense
  knowledge.
\newblock \emph{North {A}merican Chapter of the Association for Computational
  Linguistics (NAACL)}.

\bibitem[{Wang et~al.(2018)Wang, Yu, Guo, Wang, Klinger, Zhang, Chang, Tesauro,
  Zhou, and Jiang}]{wang2018r}
Shuohang Wang, Mo~Yu, Xiaoxiao Guo, Zhiguo Wang, Tim Klinger, Wei Zhang, Shiyu
  Chang, Gerry Tesauro, Bowen Zhou, and Jing Jiang. 2018.
\newblock R 3: Reinforced ranker-reader for open-domain question answering.
\newblock In \emph{AAAI Conference on Artificial Intelligence (AAAI)}.

\bibitem[{Xu and Durrett(2021)}]{xu2021dissecting}
Jiacheng Xu and Greg Durrett. 2021.
\newblock Dissecting generation modes for abstractive summarization models via
  ablation and attribution.
\newblock \emph{Association for Computational Linguistics (ACL)}.

\bibitem[{Xu et~al.(2021)Xu, Zhu, Xu, Liu, Zeng, and Huang}]{xu2021fusing}
Yichong Xu, Chenguang Zhu, Ruochen Xu, Yang Liu, Michael Zeng, and Xuedong
  Huang. 2021.
\newblock Fusing context into knowledge graph for commonsense question
  answering.
\newblock In \emph{Association for Computational Linguistics (ACL)}.

\bibitem[{Yu et~al.(2021)Yu, Zhu, Fang, Yu, Wang, Xu, Ren, Yang, and
  Zeng}]{yu2021kg}
Donghan Yu, Chenguang Zhu, Yuwei Fang, Wenhao Yu, Shuohang Wang, Yichong Xu,
  Xiang Ren, Yiming Yang, and Michael Zeng. 2021.
\newblock Kg-fid: Infusing knowledge graph in fusion-in-decoder for open-domain
  question answering.
\newblock \emph{arXiv preprint arXiv:2110.04330}.

\bibitem[{Zhang et~al.(2020)Zhang, Zhao, Saleh, and Liu}]{zhang2020pegasus}
Jingqing Zhang, Yao Zhao, Mohammad Saleh, and Peter Liu. 2020.
\newblock Pegasus: Pre-training with extracted gap-sentences for abstractive
  summarization.
\newblock In \emph{International Conference on Machine Learning (ICML)}.

\bibitem[{Zhang et~al.(2021)Zhang, Sun, Gao, Fang, Brockett, Galley, Gao, and
  Dolan}]{zhang2021joint}
Yizhe Zhang, Siqi Sun, Xiang Gao, Yuwei Fang, Chris Brockett, Michel Galley,
  Jianfeng Gao, and Bill Dolan. 2021.
\newblock Joint retrieval and generation training for grounded text generation.
\newblock \emph{arXiv preprint arXiv:2105.06597}.

\bibitem[{Zhou et~al.(2021)Zhou, Lee, Selvam, Lee, and
  Ren}]{zhou2021pretraining}
Wangchunshu Zhou, Dong-Ho Lee, Ravi~Kiran Selvam, Seyeon Lee, and Xiang Ren.
  2021.
\newblock Pre-training text-to-text transformers for concept-centric common
  sense.
\newblock In \emph{International Conference on Learning Representations
  (ICLR)}.

\end{thebibliography}
\bibliographystyle{acl_natbib}

\end{document}